\documentclass{article}

\usepackage{arxiv}

\usepackage[utf8]{inputenc} 
\usepackage[T1]{fontenc}    
\usepackage{hyperref}       
\usepackage{url}            
\usepackage{booktabs}       
\usepackage{amsfonts}       
\usepackage{nicefrac}       
\usepackage{microtype}      
\usepackage{lipsum}
\usepackage{enumitem}
\usepackage{graphicx}
\usepackage{float}

\title{Offensive language detection on Twitter}

\author{
  Nikhil Chilwant \\
  Matriculation No. : 2577689\\
  \textit{s8nichil@stud.uni-saarland.de}\\
 \\
   \And
  Syed Taqi Abbas Rizvi \\
  Matriculation No. : 2577651\\
  \textit{s8syrizv@stud.uni-saarland.de}\\
 \\
   \And
 Hassan Soliman \\
  Matriculation No. : 2576774\\
  \textit{s8hasoli@stud.uni-saarland.de}\\
}

\begin{document}
\maketitle

\begin{abstract}
Detection of offensive language in social media is one of the key challenge for social media . Researchers have proposed many advanced methods to accomplish this task. In this report, we try to use the learnings from their approach and incorporate our ideas to improve upon it. We have successfully achieved accuracy of 74\% in classifying offensive tweets. We also list upcoming challenges in the abusive content detection in the social media world.
\end{abstract}

\DeclareRobustCommand\dash{%
  \unskip\nobreak\thinspace\textemdash\allowbreak\thinspace\ignorespaces}

\section{Introduction}
The given problem is important to avoid the misuse of Twitter. We observed the given data in \emph{train.tsv}, \emph{dev.tsv} and \emph{test.tsv} and identified the following characteristics of tweets that make this problem challenging:
    \begin{enumerate}[label=(\alph*)]
        \item The short length of tweets: It is difficult to predict the sentiment, tone and context of the document with a short length of the document \cite{GoogleInc.2018}.
        \item Removal of Twitter handles, images and URLs for privacy reasons in the data limits the scope for identifying the tweet as offensive. For example, a tweet that shares a link to an offensive article/image may be classified as offensive by a human.
        \item Tweets are not present in a thread form and that limits the understanding of the context. For example, the tweet '@USER You are only believing her Because she is a Women.' is difficult to identify as offensive confidently without knowing all the tweets in the thread.
        \item It is challenging to detect a sarcastic offence. Furthermore, it is difficult to spot whether the sarcasm was meant comically or offensively. We also want to avoid classifying a funny tweet as an offensive one.
        \item No grammar policing, as discussed in the assignment.
        \item To reduce human error in the data labeling step, every training data row is usually annotated by multiple persons and usually, final decision for each label is made based on majority vote. In the case of provided \emph{train.tsv} data, it is not given whether this practice was followed or not. So removal of human errors, if present, remains to be one of the biggest challenges. The difficulty is further increased by lack of a standard definition for an offensive tweet.
 
    \end{enumerate}

\section{Literature review}
We looked at previous work done in this regard and found the following techniques and models have been used:

 \begin{enumerate}[label=(\Alph*)]
    \item Cleaning techniques:
        \begin{enumerate}[label=(\alph*)]
            \item Lower casing of words \cite{hateoffensive}.
        \end{enumerate}
    \item Classification features:
        \begin{enumerate}[label=(\alph*)]
            \item Character n-gram weighted by TF-IDF: Handles the case of spelling mistakes and spelling variations like 'kill yrslef'. It has been found to be more effective than token based n-grams \cite{schmidt2017survey, hateoffensive}.
            \item Frequency of URL mentions and punctuation comment and token lengths, capitalization, words that cannot be found in English dictionaries, and the number of non-alpha numeric characters present in tokens \cite{schmidt2017survey}.
            \item Word generalization \cite{schmidt2017survey}
            \item Sentiment analysis using PMI based approach \cite{mohammad2013nrc}, word list approach \cite{schmidt2017survey}. Further emoticons, punctuations and elongated words (e.g. 'Ohhhh')  can be used \cite{mohammad2013nrc}. 
            \item Part of Speech (POS) tags \cite{schmidt2017survey, hateoffensive}
            \item Offensiveness level score: Compute a score based on co-occurrences of offensive words in a range of distance \cite{schmidt2017survey}.
            \item Knowledge based approach: Use pre-constructed semantically connected network of words to detect offensive tweet \cite{schmidt2017survey}.
            \item Lexical feature: Use a word list to identify hate/offensive words \cite{mohammad2013nrc}.
        \end{enumerate}
    \item Classifiers:
        The supervised learning approach has been predominately used and widely suggested for classification. The Support Vector Machine (SVM) classifier performs better among classifiers \footnote{Recent papers suggest using Deep Learning approach but we ruled it out as it is out of scope for this course.}\cite{schmidt2017survey}.
\end{enumerate}

\section{Our approach}
\label{sec:headings}
We used the learnings from the literature review and tried to develop an approach on top of them. First, we finalized features and text pre-processing technique. Later, we improved the performance by using over-sampling and then analyzed our results. This section presents all carried out steps in the same sequence.

\subsection{Pre-processing}

    We did not correct or remove any row from the training data because the language data is subjective, and we did not want to introduce any biases from our side. First, we removed repetitive and uninformative words as a part of the pre-processing step. We also ignored the case of words by converting all characters to lower-case. The tweet obtained at the end of this step is called as a 'clean tweet'. However, it should be noted that, in \emph{twitterverse}, upper-cased words are called shouting.
    
    Secondly, we used NLTK's Wordnet Lemmatizer \cite{nltk_api} for lemmatizing and Porter Stemmer for stemming the tweets obtained after the first step.

\subsection{Data balancing}
    A preliminary analysis of the training data showed an imbalance in the class labels. 69.8\% of the training data was labeled as non-offensive and remaining as offensive. Since is important to train the model for both categories of tweets equally, we need to do data balancing \cite{ImbalancedDataGoogleInc}. As the provided data is small, we applied random over-sampling on the training data to handle the data imbalance problem. For this, we used \emph{imbalance-learn} \cite{imbalanceLearn} as it provides a number of re-sampling techniques commonly used in mitigating between-class imbalance in datasets.

\newpage

\subsection{Features Selection}
\textbf{Feature/model/vector/transformer?}\\
    The below table lists all the features that we considered as well as the decision about it.
    
    \begin{table}[H]
    \begin{tabular}{p{5cm}|p{3cm}|p{5cm}|}
    \hline
    Feature                                                                                                  & Selected/Rejected & Reason                                                                                                                                                                                                                                                                                                                              \\ \hline
    TF-IDF for clean tweets with character n-gram                                                            & Selected          & It provided us with a sparse matrix. Also considering the execution was completed within reasonable time, we included the whole feature set. \\\hline
    TF-IDF for lemmatized tweets with character n-grams                                                      & Rejected          & No improvement in performance. We did reach 74\% accuracy with TF-IDF for lemmatized tweet when we used word n-gram with range between 1 to 4 but the character n-gram performed gave same or better performance in every case.                                                                                                                                                                                                                                                                                                 \\\hline
    TF-IDF for stemmed tweets with character n-grams                                                         & Rejected          & No improvement in the performance was observed after GridSearchCV.                                                                                                                                                                                                                                                         \\\hline
    Text statistic like: number of Hashtags, existence of a URL link, length of a tweet, word count of tweet & Rejected          & No improvement in the performance was observed  after GridSearchCV.                                                                                                                                                                                                                                                                                                      \\\hline
    NLTK sentimental analysis                                                                                & Rejected          & No improvement in the performance. It affected the performance negatively during GridSearchCV.                                                                                                                                                                                                                                                      \\\hline
    Profanity Analysis                                                                                       & Selected          & We used \textit{profanity-check-1.0.3} and that uses combined word list of offensive words to calculate offensiveness score for the text.                                                                                                                                                                                                 \\\hline
    Word generalisation                                                                                      & Rejected          & We could not find a word list or collection of word that caters to tweet classification task.                                                                                                                                                                                                                                         \\\hline
    POS tagging using CMU TweetNLP                                                                           & Rejected          & No improvement in the performance was observed after running preliminary experiments.                                                                                                                                                                                                                                                                                                      \\\hline
    Offensive level score                                                                                    & Rejected          & No improvement in the performance. This could be because of the short length of tweets and inability to handle spelling variations.                                                                                                                                                                                                                  \\\hline
    Knowledge based approach                                                                                 & Rejected          & Out of scope for this project                                                                                                                                                                                                                                                                                                       \\ \hline
    \end{tabular}
    \end{table}

\subsection{Classifier}
   The literature review unanimously agreed upon supervised learning using SVM classifier as the best approach, and our experiments also confirmed this. Among the available classifiers in the SciKit, SGD classifier was found to be the best performing \cite{ScikitWorkingWithText}. Its hyperparameters were determined by using \emph{GridSearchCV} (CV: Cross-validation) of SciKit. \cite{ModelHyperparameterSelection}.
   We selected weight parameters for the feature union from section 3.3 using the same GridSearchCV\footnote{\href{https://gist.github.com/nikhilbchilwant/6cfd108d6077f9a589973fa92f8c90ff}{Click here for the detailed logs}}.

\section{Model parameters and result}
    This section gives quick overview of the results that we have achieved. We analysed the results in the next section.

    During the phase of hyperparameter estimation for classifier using GraphSearchCV, the accuracy score reached 76.3\%\footnote{\href{https://gist.github.com/nikhilbchilwant/6cfd108d6077f9a589973fa92f8c90ff}{Click here for the detailed logs}}. The corresponding model parameters were found as given in table 1,2:
    
    \begin{table}[H]
    \centering
    \caption{Feature union weights.}
    \begin{tabular}{|l|l|} 
    \toprule
    Feature name                                                                                 & Weight  \\ 
    \hline
    Profanity                                                                                    & 0.8     \\ 
    \hline
    TF-IDF on clean tweet with character n-gram                                                  & 1.2     \\ 
    \hline
    Sentiment                                                                                    & 0.0     \\ 
    \hline
    \begin{tabular}[c]{@{}l@{}}TF-IDF for lemmatized tweets with\\character n-grams\end{tabular} & 0.0     \\ 
    \hline
    \begin{tabular}[c]{@{}l@{}}TF-IDF for stemmed tweets with\\character n-grams\end{tabular}    & 0.0     \\
    \bottomrule
    \end{tabular}
    \end{table}

    \begin{table}[H]
    \centering
    \caption{SGD classifier parameters}
    \begin{tabular}{|l|l|} 
    \toprule
    Parameter name                                                        & Parameter value  \\ 
    \hline
    Loss function                                                         & modified huber     \\ 
    \hline
    Penalty (aka regularization term) & l2               \\ 
    \hline
    Alpha                                                                 & 0.001         \\ 
    \hline
    Random state                                                          & 69               \\ 
    \hline
    Maximum iterations                                                    & 100              \\ 
    \hline
    Stopping criterion                & none             \\
    \bottomrule
    \end{tabular}
    \end{table}

    The final summary of tweet classification is as below:
    
    \begin{table}[H]
    \centering
    \caption{Result produced by 'metrics.classification\_report'}
    \begin{tabular}{|l|l|l|l|l|} 
    \toprule
                 & Precision            & Recall               & F1-score      & Support  \\ 
    \hline
    NOT          & 0.69                 & 0.85                 & 0.76          & 500      \\ 
    \hline
    OFF          & 0.80                 & 0.63                 & 0.70          & 500      \\ 
    \hline
    Accuracy     & \multicolumn{1}{l}{} & \multicolumn{1}{l}{} & \textbf{0.74} & 1000     \\ 
    \hline
    Macro avg    & 0.75                 & 0.74                 & 0.73          & 1000     \\ 
    \hline
    Weighted avg & 0.75                 & 0.74                 & 0.73          & 1000     \\
    \bottomrule
    \end{tabular}
    \end{table}

    \par

The confusion matrix is as shown in the figure 1:
\begin{figure}[h]
\includegraphics[scale=0.35]{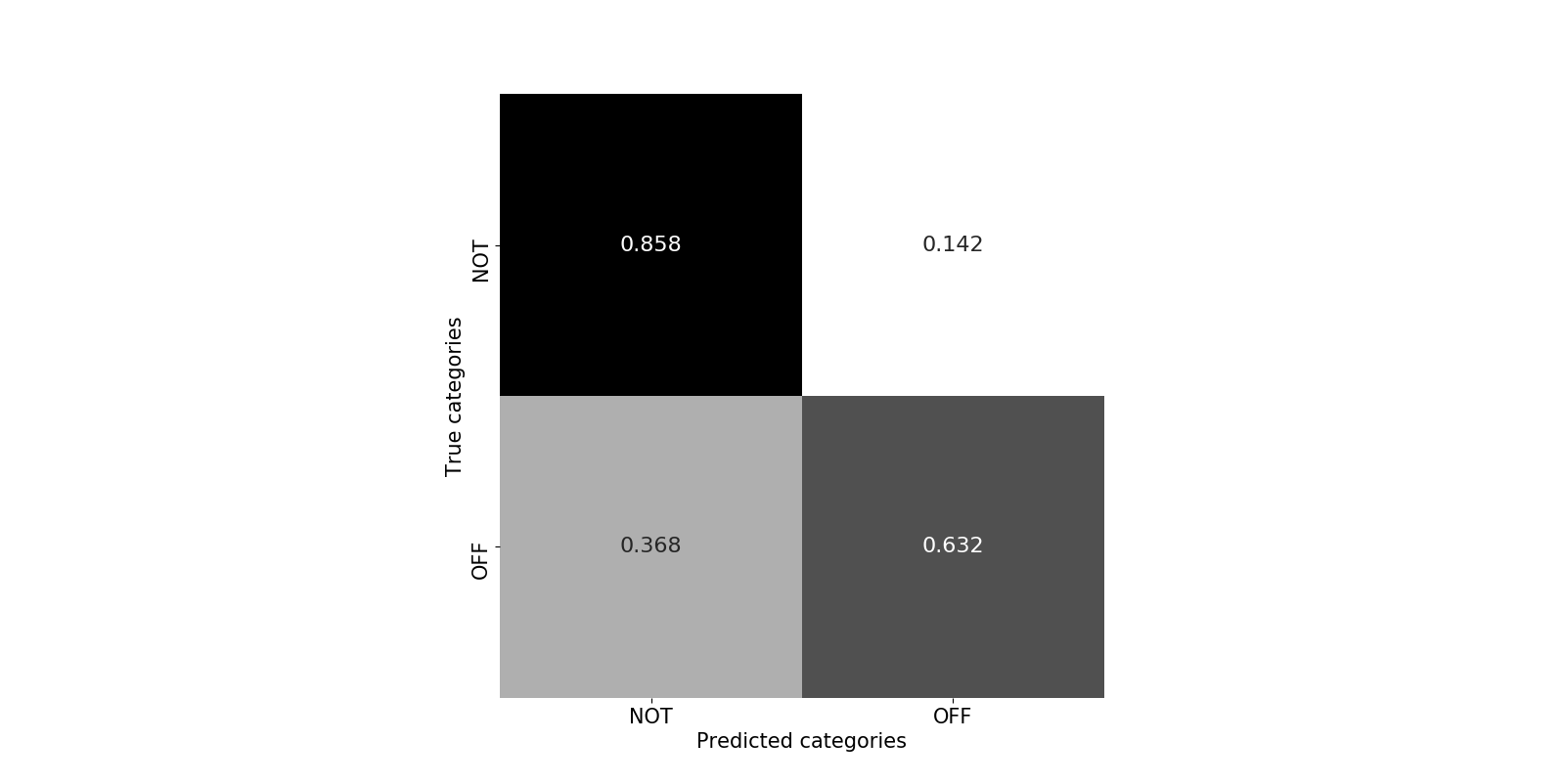}
\centering
\caption{Confusion matrix}
\end{figure}

\newpage
\section{Observations and analysis}
This section tries to explain shortcomings of the models and highlights where the model has performed good.

\subsection{Misclassification of Tweets}
    A tweet can be misclassified in two ways: an offensive tweet is classified as non-offensive and a non-offensive tweet is classified as offensive. Here are some examples of misclassified tweets from the development set which are interesting and important to understand the workings of our model. We used \textit{ELI5's} \cite{ELI5} TextExplainer functionalty for this purpose.

\subsubsection{Offensive tweets misclassified as non-offensive}
\begin{itemize}
\item \textit{@USER He is! (collision emoji)}
\newline
\textit{@USER that's unepic}
\newline
\newline
For the first tweet, after pre processing only two elements are left: \emph{'He'} and a collision emoji. Similarly, for the second tweet, after pre-processing, we get \emph{'unepic'}. These misclassifications portray the problem of lack of context. It is difficult for even a human to classify these tweets due to their short length and lack of context. Therefore, many recent studies on tweet classification consider tweets in a thread format. 
\par
The screenshorts for the ELI5 analysis gives result shown below. Note that the ELI5 treats our model like black box and then tries to find how it worked using word n-grams and so, one sees the full tweet here. The important numbers are 'mean\_KL\_divergence' and 'score'. Since the 'mean\_KL\_divergence' is close to zero, it can be inferred as the ELI5 is succeeding to guess the model. The intensity of the colour shows weights of terms\footnote{To avoid divergence from the main topic, we have limited explanation of all ELI5 terms. One can read extensive documentation at \url{ https://eli5.readthedocs.io/en/latest/_notebooks/text-explainer.html} for details about how ELI5 performs this task.}.
\begin{figure}[H]
\includegraphics[scale=0.4]{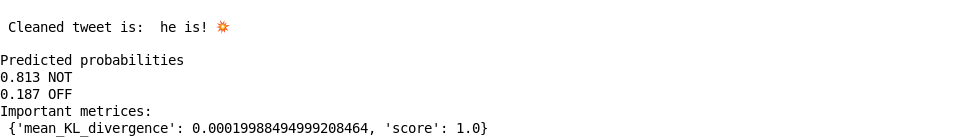}
\includegraphics[scale=0.4]{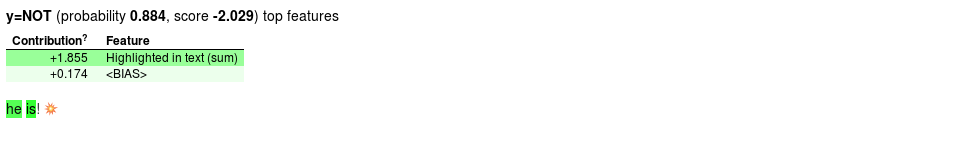}
\includegraphics[scale=0.4]{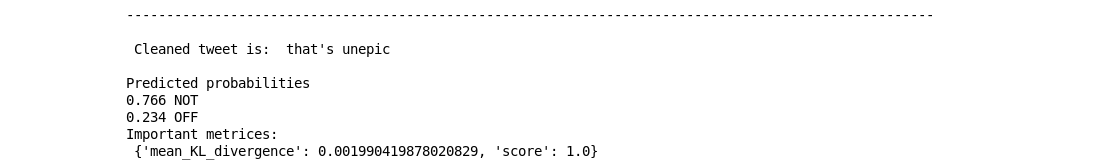}
\includegraphics[scale=0.4]{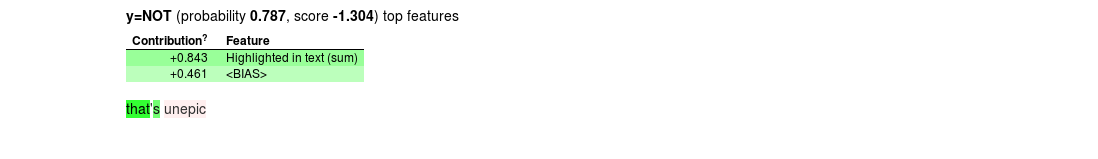}
\centering
\end{figure}

\end{itemize}
\begin{itemize}
    \item \textit{@USER @USER achichincle lamebotas!}
\newline
\newline
    Our tfidf vectors are based on character n-grams ranging from 3-grams to 6-grams with word boundary, but the training dataset is almost exclusively in English and \emph{'achichincle'} and \emph{'lamebotas'} are Mexican words. As a result, this tweet was misclassified.
    
\end{itemize}
\begin{itemize}
\item 
    \textit{@USER @USER LOVE HER!! She is a BADASS!}\\
    \textit{@USER @USER Fuuckkkk youuuuu}

    For the first one, it is a subjective choice to label it as offensive. Without the relevant context, we can not be sure if it is sarcastic or if it was labeled as such due to the word \emph{'BADASS'}. All other elements of the tweet maybe aiming to praise the subject of the tweet and with our profanity checker not tagging \emph{'BADASS'} as profanity. It was labeled as not offensive. The next tweet presents the classical problem of various spellings to write a word. The Wordnet Lemmatizer did not correctly transform \emph{'Fuuckkkk'} to \emph{'Fuck'}, hence our profanity checker did not mark it as profanity. Here we also see that our character n-grams also failed to detect the pattern.

    \begin{figure}[H]
        \includegraphics[scale=0.4]{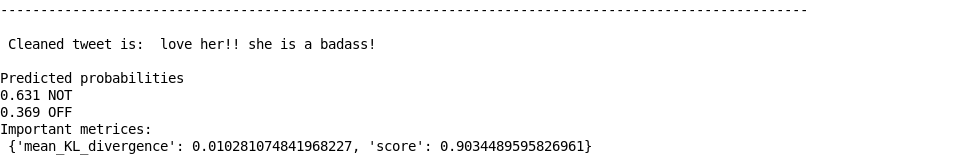}
        \includegraphics[scale=0.4]{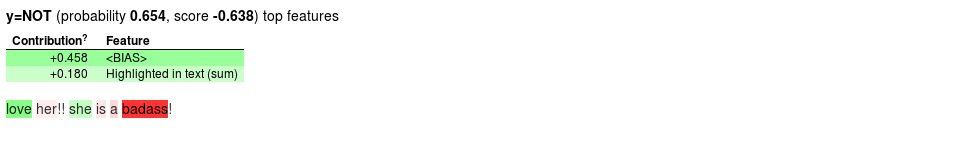}
        \includegraphics[scale=0.4]{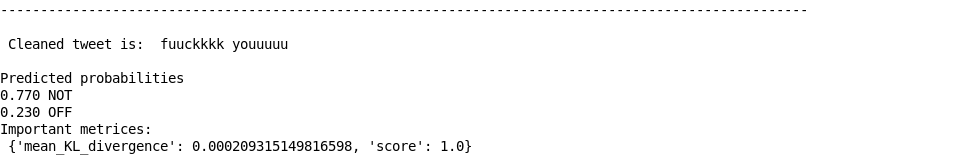}
        \includegraphics[scale=0.4]{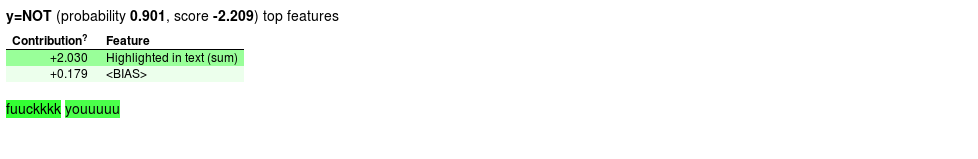}
    \centering
    \end{figure}
\end{itemize}
\subsubsection{Non-offensive tweets misclassified as offensive}
\begin{itemize}
    \item \textit{@USER She is thick though? Wtf!!}
\newline
\textit{@USER But honestly the cameras suck ass they're great if you don't zoom in}
\newline
\textit{@USER @USER hey y'all take this shit to dms aight}
\newline
\newline
These tweets were labeled as offensive by our model, which is mainly down to the use of the internet slang or curse words like \emph{'Wtf'}, \emph{'suck'}, \emph{'ass'} and \emph{'shit'} respectively. For example, the first tweet was marked as profane by the profanity checker with the confidence of 57\%. If we check the same tweet without the use of \emph{'Wtf'}; the confidence goes down to 12\%. The lack of context and the small length of the tweets also contribute to such misclassification.
    \begin{figure}[H]
        \includegraphics[scale=0.5]{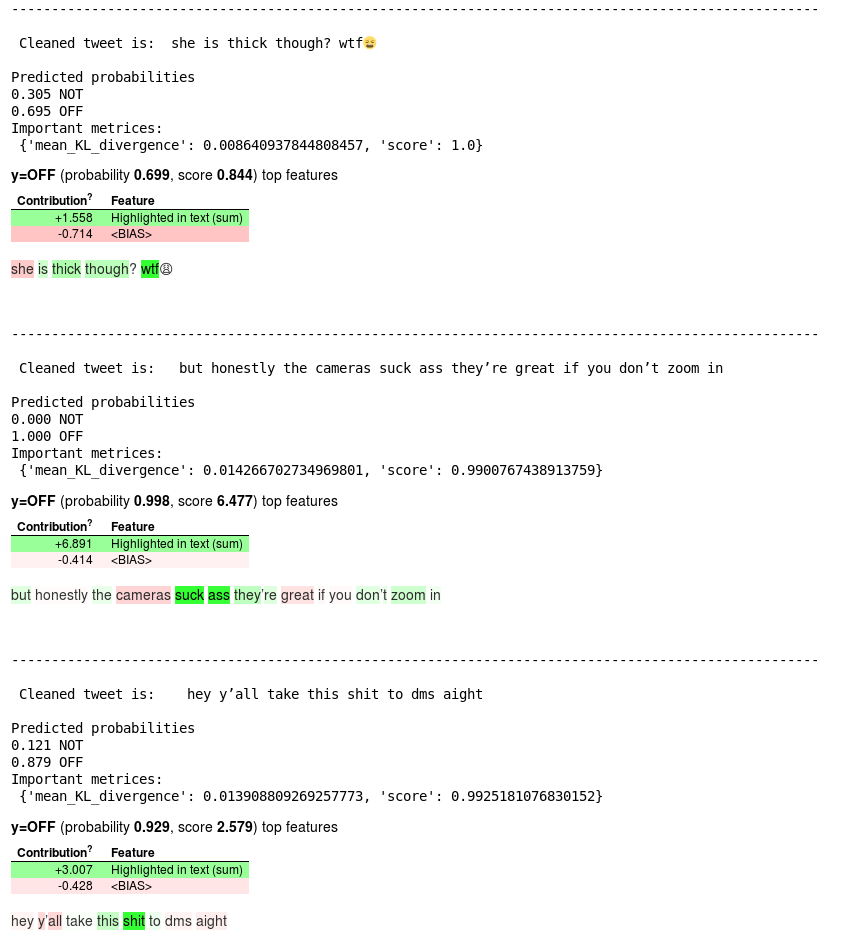}
    \centering
    \end{figure}
\end{itemize}
\begin{itemize}
    \item  \textit{@USER It would be so nice if the Trump supporters could tell us why they believe he is so great without comparing him to every criminal in the history of American politics.}
\newline
\newline
\textit{@USER @USER Our  USA WW2 soldiers were all Antifa anti-fascists too. They killed them.}
\newline
\newline
    These tweets have no profane words in the content. However the context of these tweets is quite a sensitive one according to our training dataset resulting in many offensive tweets regarding such topics. So tweets using words such as 'Antifa', 'Trump' or 'fascist', resulting them to be misclassified as offensive. It should be also noted that with character n-grams, we gain the advantage that spelling mistakes are smoothed over, however, it may find difficulty in differentiating between 'fascists' and 'anti-fascists'. There is also human judgement involved due to the polar subjects, a different dataset may label them as non-offensive.
    \begin{figure}[H]
        \includegraphics[scale=0.4]{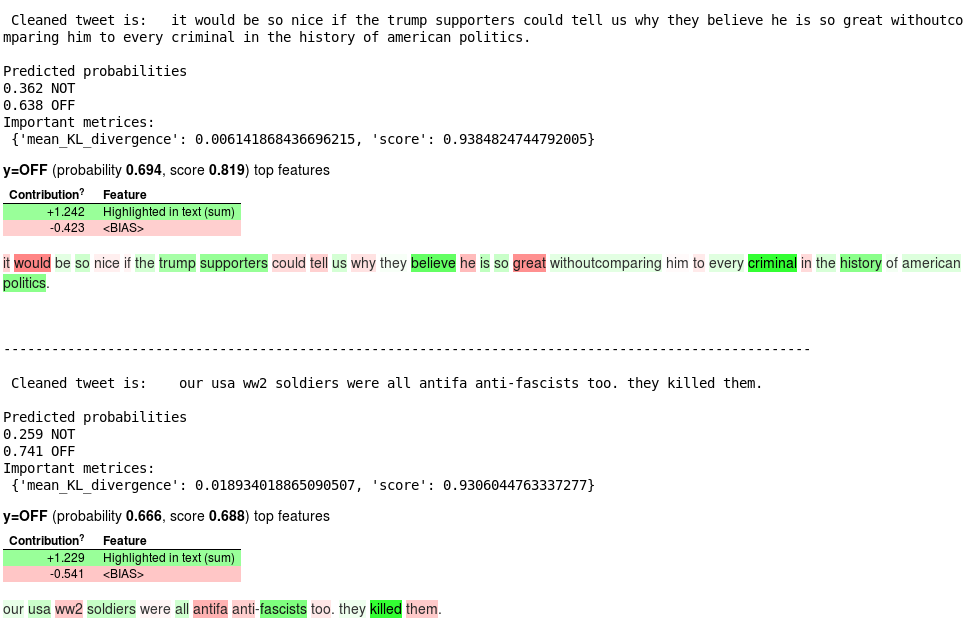}
    \centering
    \end{figure}
\end{itemize}

\subsubsection{Outlook}
The above examples explained how our model performs the classification task. Tweet is a very subjective content, with contextual relevance from outside world (outside Twitter). As a result, it is difficult to classify them. It is also heavily subject to our dataset as well. For example, the tweets related to polarizing topics were labeled as offensive in the training dataset. The topic itself became offensive in some regards according to our model. This can be mitigated by a larger and diverse dataset. Curse words are used in such creative ways, which are difficult to detect their offensive or non offensive intentions. A profanity checker maybe should accommodate for use of curse words when used without ill intention. There maybe some advanced features that could be found which makes it easier to predict the offensive nature of tweets but the one used in our model have been explained the feature selection section of our reports.

\subsection{Challenges faced in abusive content detection on social media}
This section tries to provide definition of abusive content and the challenges faced in detecting it. 
\par
We have already listed a few important challenges in the introduction. Due to subjectivity involved, no final consensus about the definition of abusive content has been reached yet \cite{hateoffensive}. For the purpose of this report, however, we define abusive content as 'a content that is aimed at making derogatory, humiliating or insulting remarks towards any person/ group based on their demography, gender, race or any other traits that could affect their psychological health in any way'. Our definition is based on a clinical basis and hence tries to reduce subjectivity. 
\par
Based on the above definition of abusive content, following are the challenges for detection of abusive content on the social media using computers:
\begin{enumerate}[label=(\alph*)]
    \item Multimedia content:\\
    Social media has multimedia content like images, GIF, videos etc. A classifier based on single media type i.e. text for the task in this assignment is an effective and easy to implement. However, as we incorporate multimedia content, we need to build more sophisticated classifiers.
    \item Multi-lingual communication:\\
    People are finding new means to communicate on the social media (e.g. using cryptic code language which only people from certain group can understand) and a lot of moderators are still catching up to this technologically. With multiple languages spilling into the internet language, new words are being adopted from various backgrounds at a higher rate than ever before.
    \item Concern for privacy:\\
    A social media enterprise has to comply to the regulations imposed by the governments but they cannot afford to risk customer's privacy for this task. However, if one has information about the producer of the content, one could make more informed guess by looking at the past content. It is an interesting algorithmic challenge to come up with an algorithm that uses the user information without giving them a feeling that the \emph{'big brother'} is watching.
    \item Context and Debate:\\
    Classifiers also have to understand to about the context of the content. A lot of abusive or offensive content may be used for educational purpose, to protest against it or to provide news on it. A system should be able to differentiate when abuse is being promoted and otherwise.
    \item Enormity of Task:\\
    With the rise of social media, huge amount of content is produced everyday. Many social media organizations are trying to develop automatic classifiers but human judgement is still needed. With rise of social media usage around the world, the definition of abusive content may have to become more nuanced than the current meaning or require regional moderation. Additionally, this rising volume translates to rising computational requirements.
\end{enumerate}

\section{Advance Analysis}
    This section lists our new ideas that can help in improve accuracy of the tweet classification. We have tried some of them and provided the source code as well.
\subsection{Word2Vec}

    We looked at Word2Vec word embedding with extracting meaningful vectors that helps not only with decreasing the dimensionality of the data, but also has shown improvement in accuracy and prediction \cite{word2Vec}.
    \newline
    \newline
    Word2Vec tries to obtain vectors that best describe the words and extract certain relations, for example words like "queen" and "king" would have similar vectors since they appear in the same window or context. 
    \newline
    \newline
    We tried to apply that to our data and test it against the same model explained previously, but the model performed worse and it resulted an accuracy of ~51\%. 
    \newline
    \newline
    This maybe explained by the training dataset of Word2Vec. It is a huge dataset compiled from different forms of formal literature like novels, newspapers, books and reports. However, Internet language has a different lexicon and many words have different connotation in \emph{twitterverse} than they would in common literature.
    \newline
    \newline
    We could not find any resource similar to Word2Vec for internet language. Here we should also note the rate of change of the common verbiage of the internet language is unusually erratic and rapid compared to formal languages. This makes having a resource like Word2Vec a difficult task.
    
    \par
    
    \subsection{Advanced Profanity Checker}
    Our model misclassified the tweet \textit{'@USER @USER Fuuckkkk youuuuu'} as our profanity checker is dictionary based approach. We propose an improvement in the profanity checker would pre-trained using a character n-gram approach. This will help us to detect spelling variations in the profane words.
    \par
    In the internet language, we see ample of curse words for myriads of uses including as adjective, adverbs or verb. Any content is not marked as offensive by the use of profanity but by intention, subject or opinion, an advance profanity checker should consider this also.
    
    \subsection{Polarizing Topics or Terms}
    When we observed results obtained after running the classifier on development data we observed that whenever words like 'Trump', 'politics', 'criminal' come, the tweet is likely to be classified as offensive as the training data had offensive tweets containing those words. This increases the chance of misclassifying a non-offensive tweet as offensive.\\ 
    \par
    Topics like politics are controversial in nature and we need to apply stronger offensive filters for tweets about them by focusing on other terms that can provide clue for classification. For example, a tweet like 'Trump is good.' should not be classified by heavily relying on a single word 'Trump' but by considering that this words indicates a controversial/multi-opinion topic and hence other terms like 'good' should not be neglected.\\
    \par
    We propose finding the polarizing topics in the training data by doing topic extraction and sentiment analysis on the content. And if the training data has similar amount of tweets for positive and negative sentiment for that class label, then it can be classified as polarizing subject. Once a tweet is identified as about polarizing subject, we downsize weights for words like 'Trump' and reclassify the tweet.
    
    \subsection{Representation Learning}
    As explained above, some words may have a heavy connotation due to the training data. For example, whenever 'fascists' was used in a tweet, the model likely predicted it was an offensive tweet. However, if there is a tweet like 'Fascists are BAD', it is likely not an offensive tweet.
    
    For this we propose, doing topic extraction and sentiment analysis on the training data and then using representation learning techniques determine the leaning of the tweet. For example, if a topic has a heavy imbalance of offensive tweets with positive sentiment, then a tweet with negative representation for that subject should be classified as non-offensive. For example, 'war is always a tragedy' should be classified as a non-offensive tweet where as 'war is a great thing' should be classified as offensive.

\newpage
\bibliographystyle{unsrt}
\bibliography{references}

\end{document}